\documentclass[letterpaper, 10 pt, conference]{ieeeconf}  % Comment this line out if you need a4paper

\usepackage{graphics} % for pdf, bitmapped graphics files
\usepackage{epsfig} % for postscript graphics files
\usepackage{mathptmx} % assumes new font selection scheme installed
\usepackage{times} % assumes new font selection scheme installed
\usepackage{amsmath} % assumes amsmath package installed
\usepackage{amssymb}  % assumes amsmath package installed
\usepackage{tabularx} %by ZW
\usepackage{graphicx} %插入图片的宏包
\usepackage{float} %设置图片浮动位置的宏包
\usepackage{subfigure} %插入多图时用子图显示的宏包
\usepackage{parskip}
\usepackage{multirow}
\usepackage{tcolorbox}
\usepackage{amsmath}
\usepackage{amssymb}
\usepackage{booktabs}
\usepackage[pagebackref,breaklinks,colorlinks]{hyperref}
\IEEEoverridecommandlockouts                              % This command is only needed if 
\title{\LARGE \bf
A Cascading Cooperative Multi-agent Framework for On-ramp Merging Control Integrating Large Language Models
}

\author{Miao Zhang$^{1*}$, Zhenlong Fang$^{2*}$, Tianyi Wang$^{3}$, Qian Zhang$^{4}$ ,Shuai Lu$^{1}$, Junfeng Jiao$^{5}$, Tianyu Shi$^{6\dag}$%
\thanks{$^{1}$Miao Zhang and Shuai Lu are with the Shenzhen International Graduate School, Tsinghua University, Shenzhen 518055, China. Email: zhangmiao@sz.tsinghua.edu.cn, shuai.lu@sz.tsinghua.edu.cn, wang.xq@sz.tsinghua.edu.cn}%
\thanks{$^{2}$Zhenlong Fang is with the Department of Computer Science \& Engineering, University of Minnesota, Twin Cities, Minneapolis, MN 55455, USA. Email: fang0540@umn.edu}%
\thanks{$^{3}$Tianyi Wang is with the Department of Mechanical Engineering \& Materials Science, Yale University, New Haven, CT 06511, USA. Email: tianyi.wang.tw727@yale.edu}%
\thanks{$^{3}$Qian Zhang is with the Computer Vision, BYD COMPANY LIMITED, Shenzhen, China. Email: zhngqn@foxmail.com}   \\
\thanks{$^{5}$Junfeng Jiao is with the School of Architecture, University of Texas at Austin, Austin, TX 78712, USA. Email: jjiao@austin.utexas.edu}%
\thanks{$^{6}$Tianyu Shi is with the Faculty of Applied Science \& Engineering, University of Toronto, Toronto, ON M5S 1A1, Canada. Email: ty.shi@mail.utoronto.ca}%
\thanks{$*$These authors contributed equally to this work.}%
\thanks{$\dag$Corresponding author}}

\begin{document}

\maketitle
\thispagestyle{empty}
\pagestyle{empty}

%%%%%%%%%%%%%%%%%%%%%%%%%%%%%%%%%%%%%%%%%%%%%%%%%%%%%%%%%%%%%%%%%%%%%%%%%%%%%%%%
\begin{abstract}

Traditional Reinforcement Learning (RL) suffers from replicating human-like behaviors, generalizing effectively in multi-agent scenarios, and overcoming inherent interpretability issues. 
These tasks are compounded when  deep environment understanding, agent coordination and dynamic optimization are required. 
While Large Language Model (LLM) enhanced methods have shown promise in generalization and interoperability, they often neglect  necessary multi-agent coordination. 
Therefore, we introduce the Cascading Cooperative Multi-agent (CCMA) framework, integrating RL for individual interactions, a fine-tuned LLM for regional cooperation, a reward function for global optimization, and the Retrieval-augmented Generation mechanism to dynamically optimize decision-making across complex driving scenarios. 
Our experiments demonstrate that the CCMA outperforms existing RL methods, demonstrating significant improvements in both micro and macro-level performance in complex driving environments.Our project page is \href{}{https://miaorain.github.io/rainrun.github.io/}

\end{abstract}

%%%%%%%%%%%%%%%%%%%%%%%%%%%%%%%%%%%%%%%%%%%%%%%%%%%%%%%%%%%%%%%%%%%%%%%%%%%%%%%%
\section{INTRODUCTION}

\label{sec:intro}
With the advancement of artificial intelligence, machine learning techniques are increasingly applied in the field of advanced systems \cite{ma2025street,  he2025enhancing1, he2025enhancing2, zhang2024adagent, zhang2024retinex, zhang2023scrnet, li2024gagent, 1_Reinforcement_learning_An_introduction}.
Reinforcement Learning (RL) stands out as a subfield that focuses on enabling agents to learn decision-making and action-taking in multi-agent environments, striving to achieve specific goals through trial-and-error learning \cite{1_Reinforcement_learning_An_introduction, Khamis2023,  Deep_Reinforcement_Learning_for_Autonomous_Driving:A_Survey}.

Extensive literature regarding Multi-agent Reinforcement Learning (MARL) explores methods to foster agent coordination \cite{oroojlooy2023review, carta2023grounding, zhu2022survey, chen2023deep,wu2023iplan}. However, these approaches encounter several challenges: 1) Sample inefficiency and lack of generalization in online learning \cite{denseSafety, comfortable_driving,Survey_LLM_AD}. 2) Difficulty in promoting general cooperative behaviors across diverse tasks. 3) Complexity in interpreting cooperative mechanisms \cite{davarynejad2011motorway,sun2024novel}.

To deal with above challenges, a promising shift is emerging with the advent of In-context Learning (ICL) applied in multi-agent systems \cite{wu2023iplan, davarynejad2011motorway, ICT, dong2022survey}, where Large Language Models (LLMs) make predictions based on contexts augmented with a few examples \cite{llm_rl_1a, llm_rl_1b, llm_rl_review}. 
Through natural language prompts to specify objectives, LLMs translate agents' intentions into actionable reward signals, intuitively guiding RL agents towards desired outcomes \cite{5_Reward_design_with_language_models,Adaptive_Reinforcement_Learning_with_LLM-augmented_Reward_Functions}. 
This innovative interaction between LLMs and RL agents not only advances automated driving technology but also highlights LLMs' versatility in complex decision-making processes \cite{cui2024receive,Driving_with_LLMs,DriveMLM,Drive_Like_A_Human}.

However, there still remains several areas for improvements: 
1) Further exploration is needed for combining RL and LLMs: The RL-based method requires numerous interactive samples and is prone to local optima. 
2) The full potential of LLMs needs to be unleashed: Current approaches in LLMs sequentially process individual information, lacking regional information, and therefore fail to utilize high-level semantic concepts.
3) Reward optimization needs to be dynamically adjusted: Although LLMs can convert agents' intentions into reward signals, the dynamic nature of environments poses challenges in adapting to varying scenarios.

The main contributions of this paper are outlined below:
\begin{itemize}
    \item \textbf{Hierarchical Multi-agent Optimization:} To improve automated driving tasks in complex scenarios, we propose a new approach dividing the multi-agent optimization problem into three levels: individual, regional, and global level. We also developed a training model, the Cascading Cooperative Multi-agent (CCMA) framework, combining RL models and LLMs to improve the three-level optimization incrementally. 
    \item  \textbf{Integration of RL and LLMs:} In the CCMA, we adopted RL's high responsiveness for individual optimization. For regional optimization, we introduced a fine-tuned LLM to utilize its spatial and advanced semantic reasoning capabilities, augmented by road network images and high-level semantic information. For global optimization, we employed the Retrieval-augmented Generation
(RAG)  strategy to iteratively enhance reward design in LLMs.
    \item \textbf{Analysis and Ablation Studies: } We conducted comparative analyses and ablation experiments to delve into key influencing factors in highway on-ramp merging scenarios, including high-level semantic information, dynamic factors, and the RAG strategy. We emphasize the critical role of cascading coordinated communication and dynamic functional alignment in achieving optimal performance in complex multi-agent tasks.
\end{itemize}

\section{Problem Formulation and Modeling}
\label{sec:problem}

%Even though previous research has explored the combination of LLMs and RL, three key questions still need further investigation.

%\begin{itemize}
%\textbf{Q1: How does our framework achieve an efficient combination of RL and LLMs for optimal results?}

%\textbf{Q2: What role do LLMs play in facilitating cooperation among vehicles under dynamic traffic conditions?   }

%\textbf{Q3: What are the key factors contributing to the overall performance of the proposed framework?}

%\end{itemize}
%To find the answer to the above question, problem formulation and modeling can be illustrated by the following content.
\subsection{Multi-agent Markov Decision Process}

An autonomous driving scenario involving multiple Connected and Automated Vehicles (CAVs) can be accurately modeled as a Multi-agent Markov Decision Process (MMDP), extending the markov decision process to accommodate multiple agents. 
The MMDP formalization is represented as:
$M = \langle S, \mathbf{A}, P, \mathbf{R}, \gamma \rangle$ . Where, 

\begin{itemize}
    \item $S$ is the state space.
    \item $\mathbf{A} = A_1 \times A_2 \times \ldots \times A_n$ is the joint action space.
    \item $P: S \times \mathbf{A} \times S' \rightarrow [0, 1]$ is the transition probability function.
    \item $\mathbf{R}: S \times \mathbf{A} \times S' \rightarrow \mathbb{R}^n$ is the joint reward function.
    \item $\gamma$ is the discount factor.
\end{itemize}

%To enhance the coordination among agents, we integrate an LLM into the reward function, using natural language processing to generate and modify the rewards based on the context of multi-agent interactions and objectives.

\subsection{Joint Reward Function}

To facilitate cooperative behaviors and maximize the overall performance, the reward function for each agent should consider both individual performance and collective outcomes. 
%To enhance safety, we introduce a penalty for excessive acceleration to discourage abrupt driving maneuvers. 
The joint reward function is defined as Equation \ref{eq:1}:
\begin{equation}
\begin{aligned}
\label{eq:1}
    R_i(s, \mathbf{a}, s') = & \ w_{\text{flow}} \cdot R_{\text{flow}}(s) \\
    + & \ w_{\text{comf}} \cdot R_{\text{comf}}(a_i) \\
    + & \ w_{\text{coop}} \cdot R_{\text{coop}}(s, \mathbf{a}, s') \\
    + & \ w_{\text{safe}} \cdot R_{\text{safe}}(s, \mathbf{a}, s')
\end{aligned}
\end{equation}

Where, $w_{\text{flow}}$, $w_{\text{comf}}$, $w_{\text{coop}}$, and $w_{\text{safe}}$ are the weights assigned to the flow, comfort, cooperation and safety component of the reward function, respectively;  $R_{\text{flow}}$, $R_{\text{comf}}$, $R_{\text{coop}}$, and $R_{\text{safe}}$ are the corresponding reward functions; and $a_i$ is the acceleration of agent $i$, and its $\text{threshold}$ is a predetermined value of acceptable maximum acceleration change beyond which penalties are incurred. 
%Where, $R_{\text{coop}}$ particularly rewards actions that enhance system-wide outcomes, such as synchronized lane changes or speed adjustments that improve overall traffic flow.

\subsection{Multi-agent Cooperative Merging Based on Large Language Models}

\begin{figure*}[h]
\centering
\includegraphics[width=0.8\textwidth]{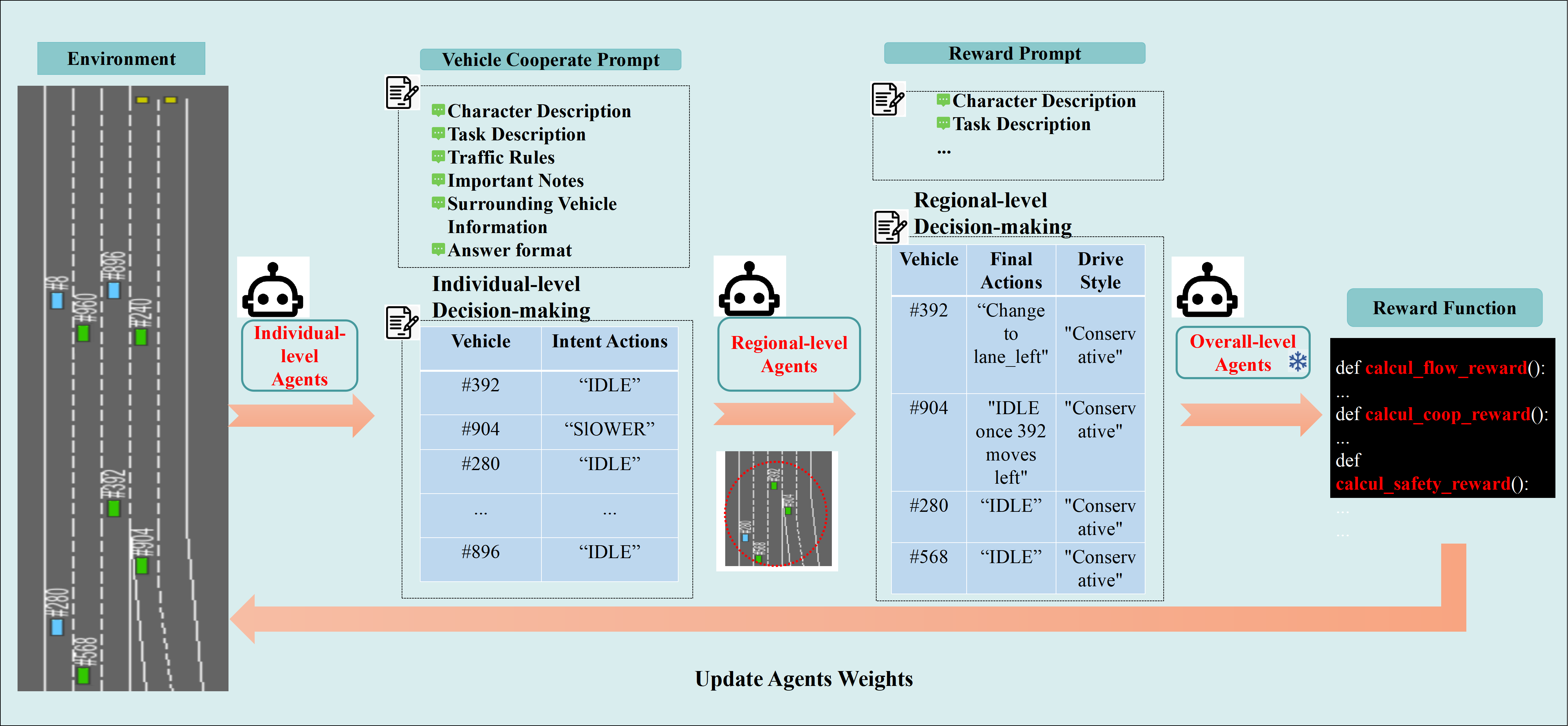}
\caption{
 The CCMA framework for cooperative merging control in a simulated traffic environment %The framework is divided into three levels of agent optimization: individual, regional, and global level. Individual-level agents are based on RL models, which input vehicle status and output each vehicle's intent actions.  Regional-level agents use the fine-tuned MMLM (GLM-4v-9B), with cooperation prompts as inputs and final actions and drive style as outputs. Global-level agents also use the GLM-4v-9B, with reward prompts as inputs and the reward functions as outputs, which can also update agents' weights.
}
\label{fig:framework}
\end{figure*}

In highway on-ramp merging areas, cooperative merging control strategies play a critical role in enhancing traffic efficiency and safety. 
Our approach leverages multi-agent systems, where each agent operates semi-independently but communicates with other agents to realize regional collaboration and global optimization. 
The main coordination framework is shown in Figure \ref{fig:framework}.

\subsubsection{Agent Architecture}
The architecture consists of three main types of agents: the Individual-level Decision-making Agent ($agent_{IND}$) by using RL models, the Region-level Decision-Making Agent ($Agent_{RDM}$) with a fine-tuned LLM, and the Global-level Decision-making Agent ($Agent_{GDM}$)  utilizing GLM-4v-9B. 
$agent_{IND}$ utilizes a pre-trained RL model to identify and avoid hazardous actions.
$Agent_{RDM}$ leverages the fine-tuned GLM-4 to optimize actions for multiple vehicles within a regional area, considering varying driving styles. 
$Agent_{GDM}$ aims to refine the reward function to evaluate $Agent_{RDM}$'s final actions based on different driving styles and dynamic traffic densities.%, considering metrics such as safety, cooperation, flow, and comfort,  

\subsubsection{Local Observations and Decision-making}
Each agent has access to local observations of immediate traffic conditions, depicted by circles in Figure \ref{fig:framework}. 
These observations form the basis for calculating both ego and cooperative rewards, which drive the agent's decision-making. 
The total rewards can be mathematically expressed as Equation \ref{eq:2}:
\begin{equation}
\label{eq:2}
R_{\text{total}, i} = w_{\text{ego}} \cdot R_{\text{ego}, i} + w_{\text{coop}} \cdot R_{\text{coop}, i}
\end{equation}
Where, $w_{\text{ego}}$ is the weight for ego component of the reward function; $R_{\text{ego}, i}$ focuses on the agent's individual performance metrics such as speed maintenance and safe following distances; and $R_{\text{coop}, i}$ is based on the agent's ability to contribute to overall traffic efficiency and safety.

\subsection{Negotiation and Local Optimization}
Agents engage in continuous negotiation to dynamically adjust their actions in response to the evolving traffic conditions. 
This process enables local optimization, allowing agents to refine their strategies for both individual benefits and overall system performance. 
The merging process, in particular, requires careful collaboration between main lane vehicles and ramp vehicles to avoid disruptions in traffic flow. 
The negotiation objective for each agent $i$ can be formulated as Equation \ref{eq:3}:
\begin{equation}
\label{eq:3}
\max_{a_i} \left\{ R_{\text{total}, i}(s, a_i, s') \right\}
\end{equation}
Where, the game is subject to traffic dynamics constraints and safety regulations.
\section{Proposed Approach}
\label{sec:approach}

%In this work, we propose a robust framework for cooperative on-ramp merging control in multi-agent systems using LLMs, as depicted in Figure \ref{fig:framework}. 
%Our approach innovatively combines RL with LLMs to enhance communication and decision-making processes among agents in dynamic traffic environments.

\subsection{Individual-level Model}
We model the multi-agent system as a set of CAVs \( \{a_1, a_2, \dots, a_N\} \), each equipped with an LLM-based controller. 
The state of each agent \( a_i \) at time \( t \) is represented by \( s_t^i \), which includes both the physical states and observable environmental factors.

\begin{figure}[h]
\centering
\includegraphics[width=0.48\textwidth]{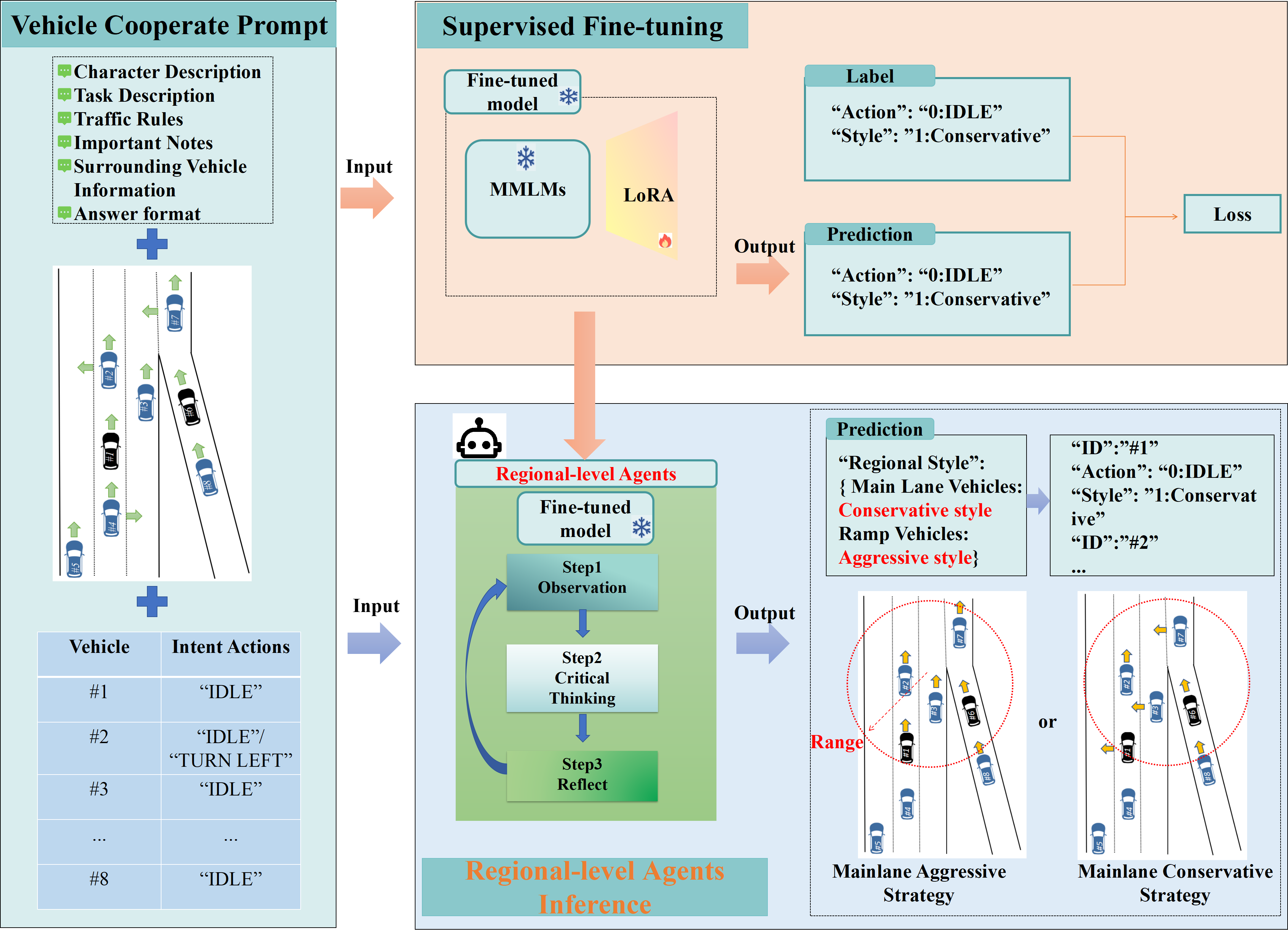}
\caption{
Regional-level coordination in the CMAA framework
%Cooperative merging control involves supervised fine-tuning and inference for regional-level agents. The process begins with a set of inputs guiding individual vehicles' initial intent actions. A fine-tuned model, enhanced by MMLM and LoRA, is trained using labeled data and deployed to make predictions for regional-level agents. The agents observe their environment, assess situations, and reflect on past decisions to determine which of the main lane vehicles and ramp vehicles have the right-of-way at the merge point. Vehicles with the right-of-way adopt an aggressive strategy, while the other vehicles take a conservative approach.
}
\label{fig:regional}
\end{figure}

\subsection{Regional-level Coordination}
Each agent's decision-making is guided by a policy \( \pi \) that maps states to a probability distribution over possible actions, modeled as Equation \ref{eq:4}:
\begin{equation}
\label{eq:4}
\pi(a_t^i | s_t^i) = \text{softmax}(\text{LLM}(s_t^i, \theta)),
\end{equation}
Where, \( \theta \) denotes the parameters of the LLM, and \( \text{LLM}(s_t^i, \theta) \) outputs the logits for each potential action \( a_t^i \). 
%The softmax function ensures that the output forms a valid probability distribution across actions.
\subsubsection{Decision-making}
%LLMs typically process inputs natural language and images rather than raw vectorized data, as shown in Figure \ref{fig:framework}. 
%Therefore, creating effective prompts that accurately describe current observations in natural language is crucial. 
%Previous research focused on leveraging LLMs' deep reasoning capabilities through innovative prompt design. 
We develop prompts based on self-reflection and Chain of Thought (CoT). 
%Section APPENDIX illustrates an example of our input prompts. 
The model receives images as input and is prompted to provide a role description, task description, tool description, and an anomaly category, with more details showing in Figure \ref{fig:regional}.
\subsubsection{Fine-tuning}
We employ GLM-4v-9B, an open-source foundational language model, as the pre-trained model. 
For parameter-efficient fine-tuning, we utilize the Low-rank Adaptation (LoRA) method, which keeps the pre-trained model weights static and integrates trainable rank decomposition matrices into each Transformer layer. 
The raw dataset, derived from a trained RL model, is adjusted for accuracy. 
%Notably, the ground truth encompasses labels for multiple vehicles, not just for a single one.
\subsubsection{Coordination}
To facilitate effective coordination, agents communicate through a structured language protocol generated by the LLM. 
The communication vector \( m_t^i \) for agent \( i \) is generated based on its current state and intended action, as shown in Equation \ref{eq:5}:
\begin{equation}
\label{eq:5}
m_t^i = \text{LLM}(s_t^i, a_t^i, \theta).
\end{equation}
Where, agents broadcast \( m_t^i \) to their neighbors, enhancing collaborative decision-making through shared information.

\subsection{Global-level Optimization}
\begin{figure*}[h]
\centering
\includegraphics[width=0.75\textwidth]{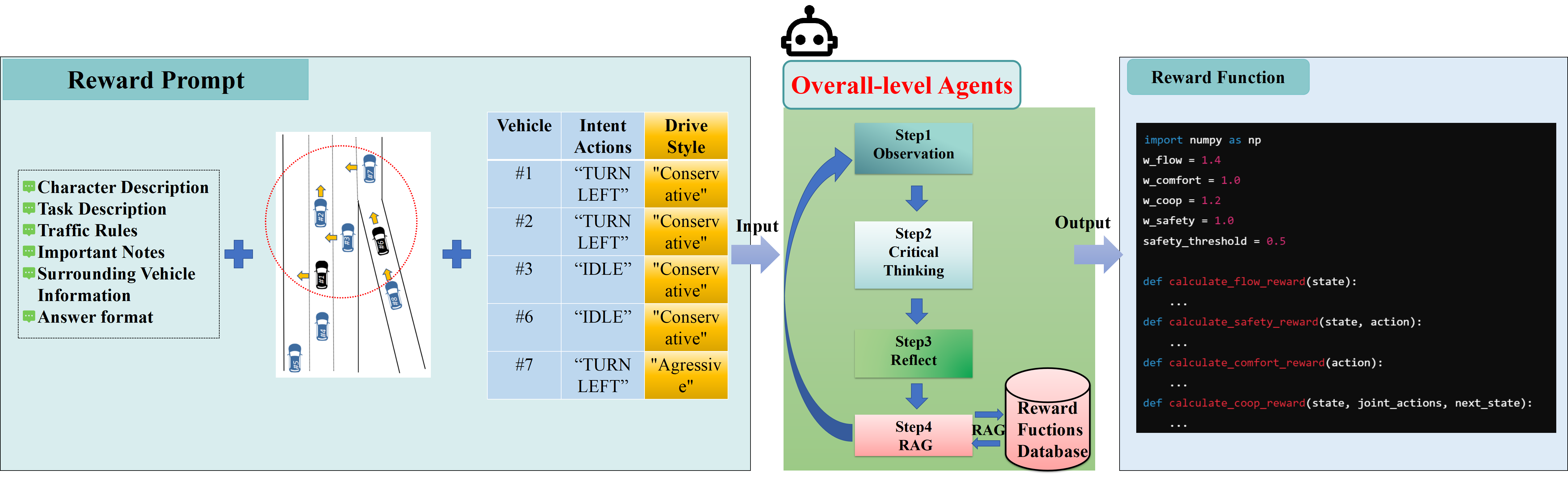}
\caption{
Global-level optimization in the CMAA framework
%This figure depicts global-level agents' decision-making in a cooperative merging control framework. It begins with initial inputs influencing intent actions and driving styles. The process involves four steps: observation, critical thinking, reflection, and RAG. The outputs update the reward functions database to fine-tune decision-making.
}
\label{fig:overall}
\end{figure*}

The global optimization process leverages the information of local optimization and the general reasoning capabilities of GLM-4v-9B to autonomously construct reward functions that balance various performance metrics. 
This study introduces an innovative strategy called "Critical Reflect CoT: Observation, Critical Thinking, Reflection, and RAG" aimed at enhancing reasoning efficiency and generating globally optimized reward functions, as shown in Figure \ref{fig:overall}. %For details on the reward designing prompt, refer to Appendix B.

% \subsection{Framework Training}
% \subsubsection{Learning Objective}
% The learning objective is to maximize the collective reward function, which integrates individual agent rewards and overall performance metrics. 
% The reward function \( R \) for agent \( i \) is defined as Equation \ref{eq:6}:
% \begin{equation}
% \label{eq:6}
% R(s_t^i, a_t^i) = r_t^i + \gamma \sum_{j \neq i} \lambda_{ij} r_t^j,
% \end{equation}
% Where, \( r_t^i \) is the immediate reward received by agent \( i \) at time \( t \) ; \( \gamma \) is a discount factor; and \( \lambda_{ij} \) weights the importance of the rewards received by surrounding agents \( j \).

\section{Experiments}
\label{sec:experiments}

%This section presents the experimental evaluation of our proposed multi-agent system for cooperative merging using LLMs. 
%We assess the system's performance in the HighWay Env, a simulated multi-lane highway environment, where multiple CAVs coordinate merging maneuvers under varying traffic conditions. 

\subsection{Dataset Processing and Simulation Setup}
Our experiments utilized a modified version of the highway traffic simulation tool~\footnote{https://github.com/Farama-Foundation/HighwayEnv} to incorporate LLM-based decision-making. 
%We simulated a multi-lane highway on-ramp merging area where ramp vehicles need to cooperatively merge into the main lanes. 
%Each CAV in the simulation is controlled by an agent using our LLM-based framework. 
Initially, simulation data were gathered, including screenshots from the HighWay Env simulator and vehicle details generated by a well-trained RL model. 
%These data were then manually validated and labeled with "style" tags (aggressive or conservative). 
To enhance LLM's multi-vehicle prediction capabilities, $i$ vehicles were randomly selected from a total of $n$ as input prompts, while the remaining $n$$-$$i$ vehicles served as prediction labels. 
The dataset comprises approximately 50,000 samples, derived from six different road maps and simulated around 600 times each, with three different traffic densities and random spawn points, capturing about 10-20 timestamps per simulation.
GLM-4v-9B was utilized as a baseline model, with each linear layer fine-tuned using LoRA. 
All models were implemented in PyTorch and trained on two A100 GPUs. 

\subsection{Baselines}
In this subsection, we compare the proposed method with several state-of-the-art MARL benchmarks, including the Multi-agent Advantage Actor-critic (MAA2C)\cite{papoudakis2020benchmarking}, the Multi-agent Proximal Policy Optimization (MAPPO)\cite{yu2022surprising}, and the Multi-agent Advantage Actor-critic with Kronecker-factored Trust Region (MAACKTR)\cite{wu2017scalable}.
All the baselines utilize parameters sharing among agents to handle dynamic agent counts, global rewards and discrete action spaces.
%\begin{itemize}
%MAA2C is an extension of Advantage Actor-critic (A2C) algorithm for multi-agent environments \cite{papoudakis2020benchmarking}. It employs a centralized critic to estimate the value function and decentralized actors for action selection, enabling effective coordination among agents.
    
%MAPPO is an extension of Proximal Policy Optimization (PPO) algorithm tailored for multi-agent systems \cite{yu2022surprising}. It features centralized training with decentralized execution, sharing policy parameters among agents and optimizing a clipped surrogate objective to ensure stable learning.
    
%MAACKTR combines MAA2C with Kronecker-factored Trust Region (K-FAC) optimization method to accelerate learning \cite{wu2017scalable}. K-FAC approximates the natural gradient, leading to more efficient policy updates and improved performance in complex multi-agent environments.
    
%\end{itemize}

\subsection{Experimental Tasks and Scenarios}
In this subsection, we designed three comprehensive scenarios to explore the integration of RL and LLMs in multi-agent autonomous driving systems, each targeting one specific research shortcoming in Section \ref{sec:intro}.

%Accident frequency in CAVs is high due to their inability to quickly respond to the complex driving environment, especially when sharing the road with Human-driven Vehicles (HDVs) \cite{chen2023deep}. 
%To better reflect the impact of CAVs on the mixed traffic, we set the CAV penetration rate at 50\% in this paper.
%Consequently, simulations in this paper were conducted with CAVs and HDVs under three different traffic densities, which showed three modes in the control range: easy mode (2-3 CAVs controlled by our model and 2-4 HDVs controlled by the HighWay Env simulator), medium mode (3-5 CAVs and 2-5 HDVs), and hard mode (4-7 CAVs and 2-6 HDVs).  Each scenario was simulated 100 times for statistical significance.

\subsubsection{\textbf{Scenario 1 Integration of RL and LLMs}}
\label{sec:scenario1}

To evaluate how our framework optimizes the integration of RL and LLMs, we designed experiments at three levels: individual, regional, and global-level optimization. 
%This setup highlights the distinct advantages of the proposed CCMA framework over other traditional MARL methods.

\begin{figure}[h]
    \centering
    \includegraphics[width=0.4\textwidth]{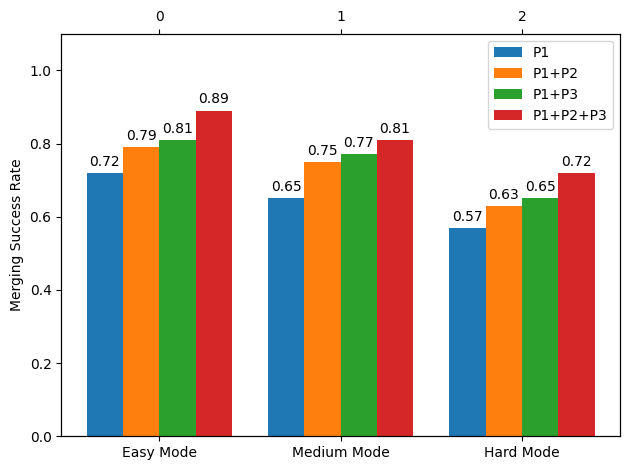}
    \caption{Comparison of merging success rates across varying traffic modes}
    \label{fig:scenario1_results}
\end{figure}

\paragraph{Individual-level Optimization}
We leverage RL's high responsiveness to swiftly process decision-making information for each vehicle. 
%The RL model is trained to handle single-vehicle scenarios, focusing on collision avoidance and real-time inferences. 
%This ensures that each vehicle can operate independently with minimal delays and immediate responses to dynamic traffic conditions.
As shown in Figure \ref{fig:scenario1_results}, merging success rates at individual-level (P1) are 0.72 in easy mode, 0.65 in medium mode, and 0.57 in hard mode, demonstrating RL's effectiveness in quick and safe individual decision-making.

\paragraph{Regional-level Optimization}
We utilize a fine-tuned Visual Language Model (VLM) to incorporate spatial and high-level semantic reasoning. 
%This model processes inputs (e.g., road network images and detailed semantic information, including driving regulations, driving styles, and the concepts of the main lane and the ramp) to enhance traffic flow and decision-making interpretability.
According to Figure \ref{fig:scenario1_results}, merging success rates at regional-level (P1 + P2) are 0.79 in easy mode, 0.75 in medium mode, and 0.63 in hard mode. 
%This indicates that regional optimization, including spatial and high-level semantic reasoning, significantly enhances the CCMA's ability to manage traffic flow efficiently and boosts better overall performance.

\paragraph{Global-level Optimization}
The framework dynamically updates reward designs based on evolving road conditions. 
These updated rewards are stored in a database and utilized by the RAG mechanism to guide data-driven decisions. 
%This approach aims to balance the trade-offs between multiple indicators such as safety, efficiency, and comfort, thereby enhancing the overall performance in dynamic environments.
As shown in Figure \ref{fig:scenario1_results}, the global-level optimization (P1 + P2 + P3) achieves the highest merging success rates: 0.89 in easy mode, 0.81 in medium mode, and 0.72 in hard mode. 
%By dynamically adjusting rewards based on real-time road network conditions, this approach effectively balances safety, efficiency, and comfort, resulting in superior performance across all traffic densities.

\begin{figure}[h]
    \centering
    \includegraphics[width=0.45\textwidth]{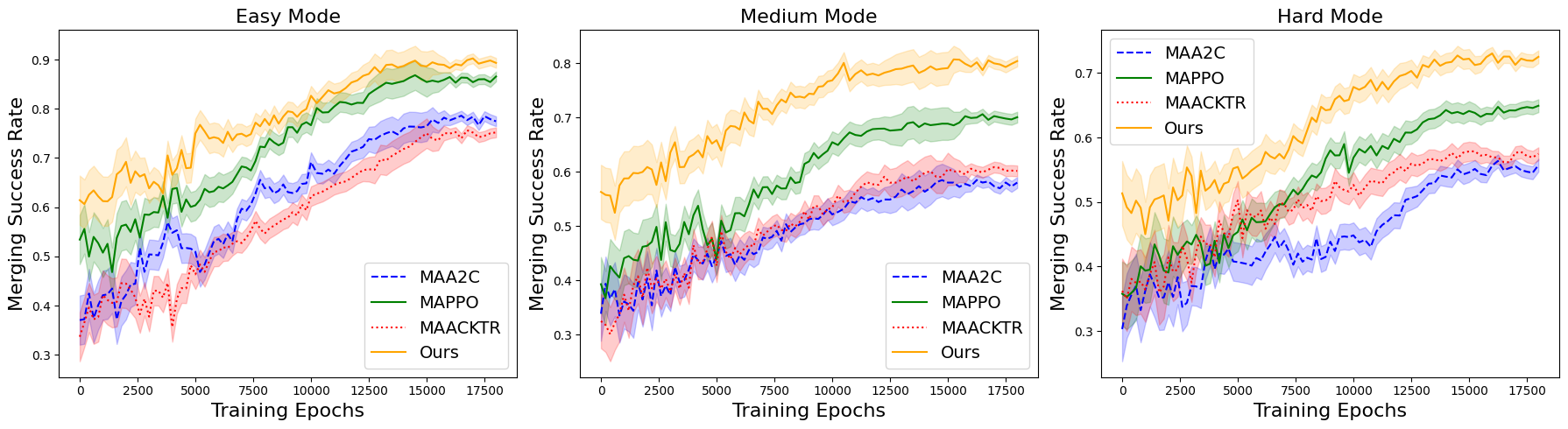}
    \caption{Comparison of merging success rates across MARL benchmarks }
    \label{fig:merging_success_rate}
\end{figure}

\paragraph{Conclusion}
Figure \ref{fig:merging_success_rate}  demonstrates that the CCMA achieves higher merging success rates across all traffic modes, particularly 0.81 in medium mode and 0.72 in hard mode, which are much higher than other MARL algorithms. 
This suggests that the LLM enhanced reward functions and communication protocols effectively promotes collaborative actions among agents, especially in dense traffic conditions.These experiments verify that our CCMA method demonstrates significant improvements in cooperative behaviors, which provides strong evidence for the potential of LLM integrated MARL approaches in autonomous driving. 
%By integrating RL and LLMs, our CCMA method demonstrates significant improvements in both cooperative behaviors and traffic efficiency, advancing the development of intelligent transportation systems.

\subsubsection{\textbf{Scenario 2 Role of LLMs in Facilitating Communication and Decision-making}}
\label{sec:scenario2}

To investigate the role of LLMs in facilitating communication and decision-making, we designed a case study on CAVs' merging process on a multi-lane highway scenario. 
%This case study aims to highlight how LLMs can dynamically adjust vehicle strategies (aggressive or conservative) based on real-time traffic conditions and the intentions of surrounding vehicles to optimize the overall system performance.

\begin{figure}[htpb]
    \centering
    \includegraphics[width=0.45\textwidth]{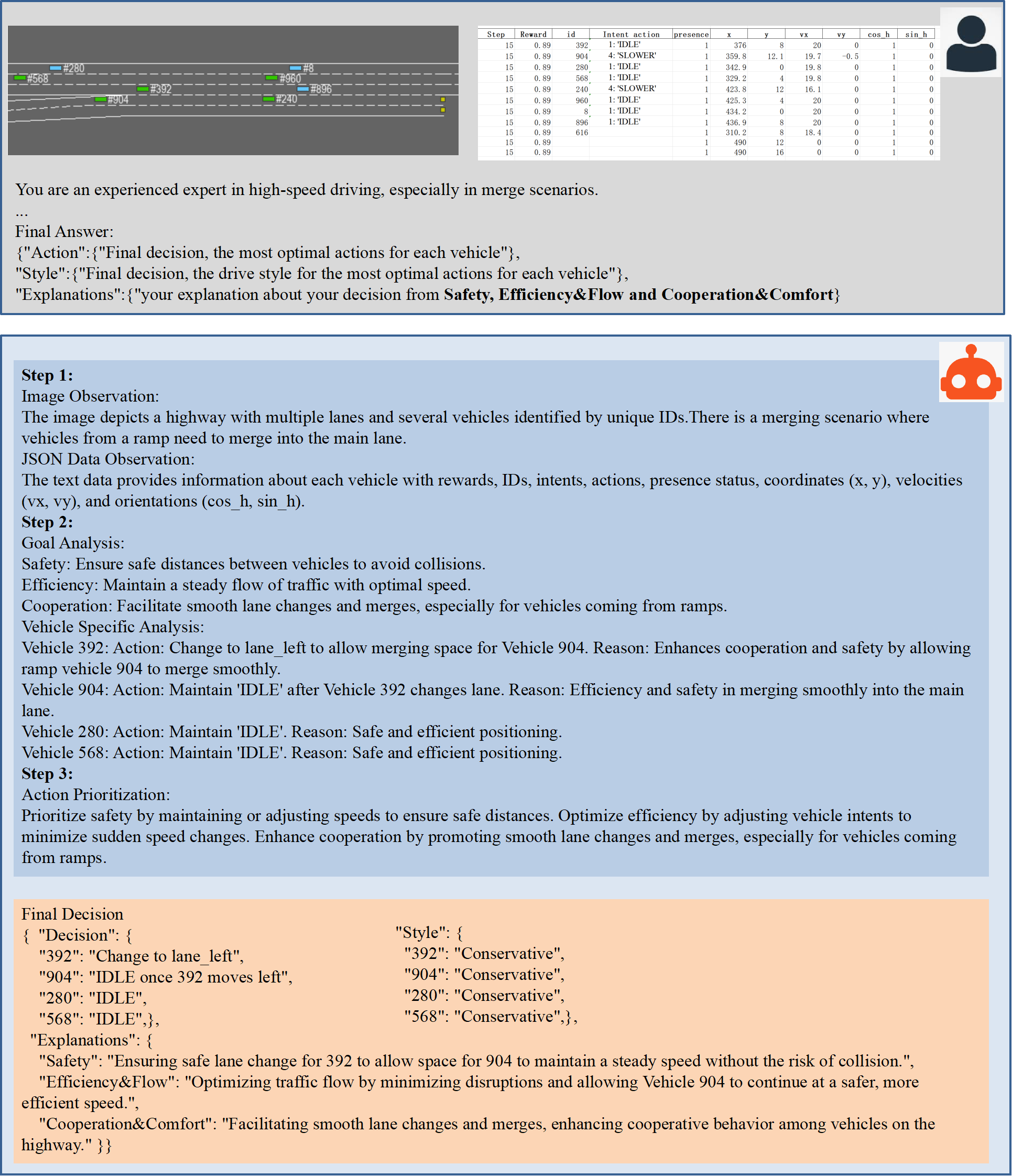}
    \caption{Vehicle intentions and actions during decision-making}
    \label{fig:case2}
\end{figure}

\paragraph{Case Study Overview}
The case study is divided into three main steps: observation, action analysis, and critical thinking, as shown in Figure \ref{fig:case2}. 
%Each step is crucial for understanding how LLMs can enhance decision-making in real-time traffic scenarios.

Step 1: Observation

% \textbf{Image Analysis}

The first step involves gathering detailed information about the highway environment. 
%The image in Figure \ref{fig:case2} depicts the multi-lane highway with multiple vehicles. 
In Figure \ref{fig:case2}, the ego vehicle (\#904) and its surrounding vehicles are analyzed. 
The accompanying JSON data provides comprehensive state information for each vehicle, including position, speed, and heading.
This step focuses on collecting all relevant data about the vehicles and the surroundings without making any decisions.
% \textbf{Detailed JSON Data Analysis}

%\begin{itemize}
%\textbf{Vehicle IDs:} Unique identifiers for each vehicle such as \#368, \#584, \#240, etc.
%\textbf{Features:} Position (x, y), velocity (vx, vy), heading (cos\_h, sin\_h), presence indicator, and reward information.
%\textbf{Example:} Vehicle \#368 has a position of (376.0, 8.0) and a velocity of (0.0, 0.0).
%\end{itemize}

Step 2: Action Analysis

% \textbf{Safety and Efficiency Goals}

The available actions for the ego vehicle (\#904) are analyzed to maintain safe distances from neighbors and to optimize traffic flow by avoiding sudden braking or acceleration. The JSON data provides a detailed framework for understanding the intent actions of each vehicle.

% \textbf{Vehicle-specific Analysis}

Each vehicle's actions are analyzed to facilitate smooth merging and lane-changing:
\begin{itemize}
    \item \textbf{Ego Vehicle (\#904):} Adjust speed to merge smoothly into the main lane.
    \item \textbf{Surrounding Vehicles:} Maintain safe distances and create gaps for the ego vehicle to merge.
\end{itemize}

Step 3: Critical Thinking

% \textbf{Prioritizing Actions}

The available actions for the ego vehicle include changing lanes, accelerating, braking, or maintaining speed. 
The final decision section in Figure \ref{fig:case2} provides a JSON formatted decision output to ensure both safety and efficiency during the merging and lane-changing process:
%These actions are prioritized to ensure optimal speed and safe distance on the main lane while handling the complexities of merging maneuvers.

% \textbf{Decision-making Process}
\begin{itemize}
    \item \textbf{Action 1:} Vehicle \#904 adjusts speed to merge smoothly into the main lane.
    \item \textbf{Action 2:} Vehicle \#392 
    Change to the left lane to allow merging space for the ego vehicle to merge.
\end{itemize}

% \textbf{Detailed Reward Structure}

%Implementation

%The Python code provided details the implementation of the reward function. 
%This code dynamically calculates the total reward for each action based on the current traffic state.

\paragraph{Conclusion}

%The case study demonstrates how LLMs can facilitate real-time communication and decision-making in autonomous driving scenarios. 
By dynamically adjusting vehicle strategies based on real-time traffic conditions and the intents of the surrounding vehicles, LLMs can significantly enhance safety, efficiency, and cooperation among CAVs. 
This LLM-RL integration leads to more optimized and human-like driving behaviors, ultimately improving the overall performance of the multi-agent system.

%\begin{figure*}[htpb]
  %  \centering
    %\includegraphics[width=0.8\textwidth]{case3.png}
 %   \caption{Vehicle intentions and actions during decision-making}
   % \label{fig:case2}
%\end{figure*}

%\begin{figure*}[htpb]
  %  \centering
    %\includegraphics[width=0.8\textwidth]{reward_output.png}
 %   \caption{Dynamic LLM integrated reward function design }
   % \label{fig:reward_design}
%\end{figure*}

\subsubsection{\textbf{Scenario 3: Key Factors of Overall Performance}}
\label{sec:scenario3}

To understand the key factors contributing to the overall performance of our CCMA framework, we investigated three key aspects: the reward function parameters, the RAG mechanism, and the choice of LLMs. 

\paragraph{Reward Function Parameters}

\begin{figure}[htbp]
    \centering
    \includegraphics[width=0.4\textwidth]{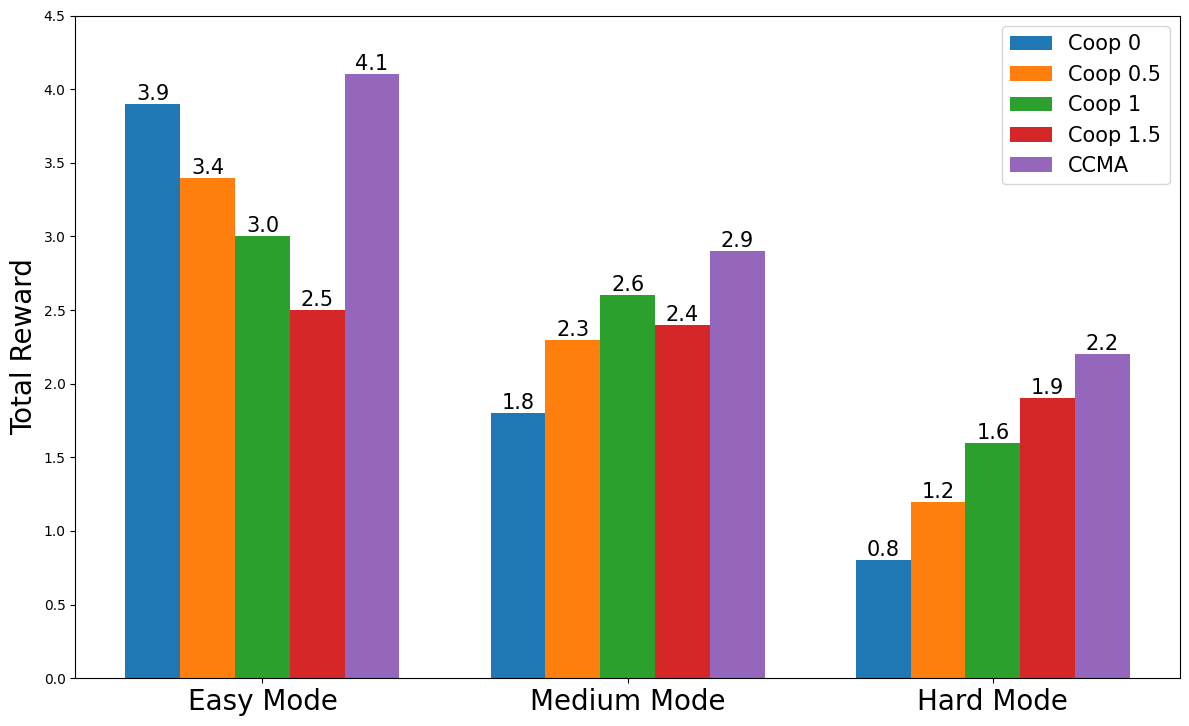}
    \caption{Overall performance across varying cooperative factors}
    \label{fig:cooperate_factor_performance}
\end{figure}

The reward function given by Equation \ref{eq:1} is dynamically adjusted to balance traffic flow, cooperation, comfort, and safety. 
The use of LLMs allows for real-time tuning of these parameters, ensuring that the agents maintain optimal performance under varying traffic conditions. 
%The reward function is given by Equation \ref{eq:1}. 
As Figure \ref{fig:cooperate_factor_performance} shows,  this dynamic adjustment of the cooperation factors allows the framework to maintain optimal performance under varying traffic conditions, reducing delays and improving the overall traffic flow.

\paragraph{Retrieval-augmented Generation Mechanism}

\begin{figure}[h]
\centering
\includegraphics[width=0.4\textwidth]{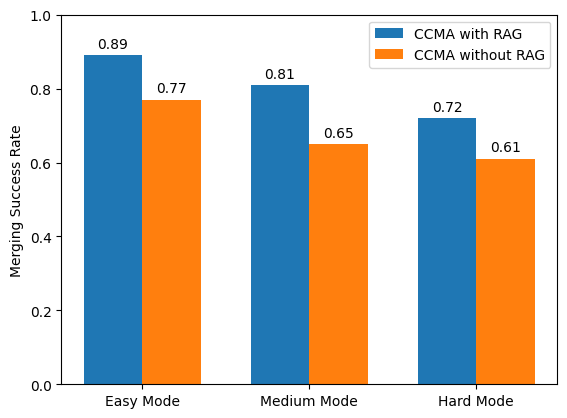}
\caption{
Impact of the RAG mechanism on merging success rates
%This chart compares the merging success rates of the CCMA framework with and without the RAG mechanism across different traffic modes.
}
\label{fig:rag_performance}
\end{figure}

The RAG mechanism enhances decision-making by retrieving relevant data from a database and generating informed decisions based on real-time road conditions. 
This mechanism allows the CCMA to dynamically update rewards and optimize the trade-offs among multiple performance indicators. 
As can be seen in Figure \ref{fig:rag_performance}, the CCMA framework with RAG achieves higher merging success rates across all traffic modes compared to the CCMA without RAG. 

%Figure \ref{fig:rag_performance} illustrates the impact of the RAG mechanism on merging success rates. 

%For example, in medium mode, the merging success rate is 0.81 with RAG, compared to 0.65 without RAG. 
%This demonstrates that the RAG mechanism significantly enhances our framework's ability to make informed decisions and adapt to dynamic environments, leading to improved performance.

\paragraph{Large Language Models}

\begin{figure}[htbp]
\centering
\includegraphics[width=0.4\textwidth]{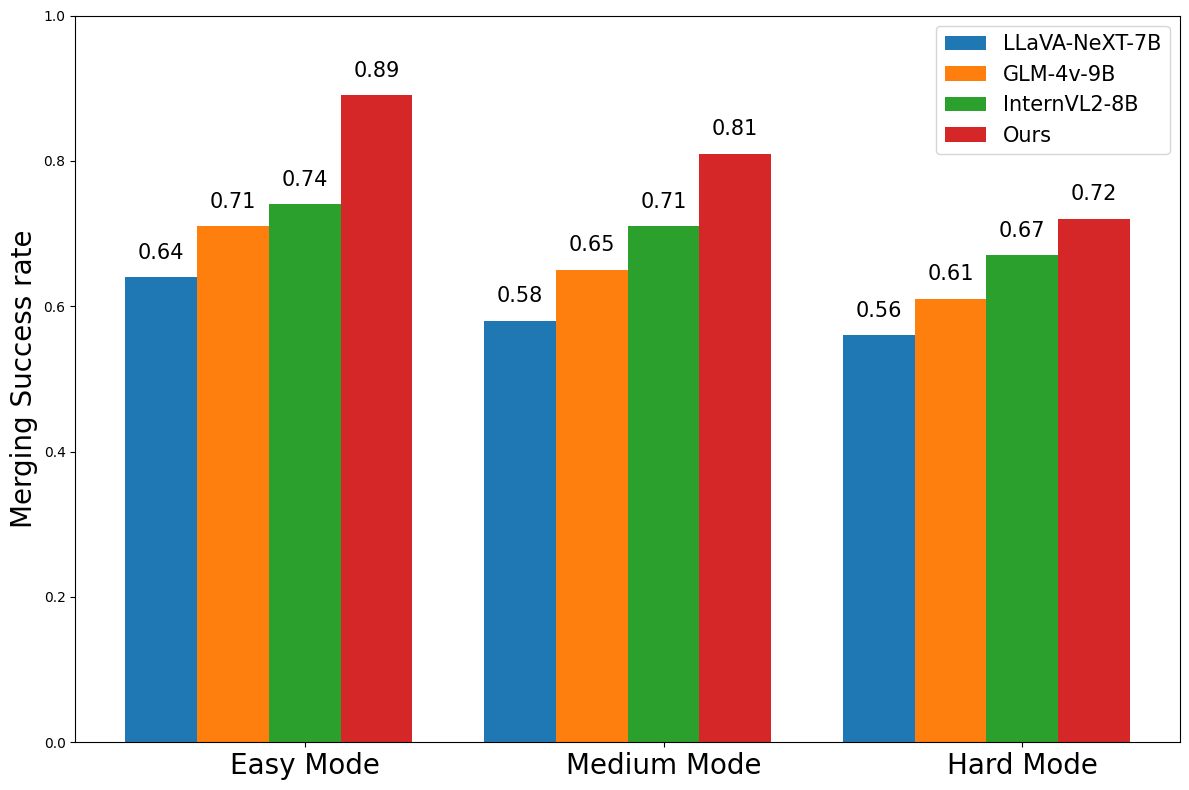}
\caption{Comparison of merging success rates across various LLMs}
\label{fig:model_kernel_performance}
\end{figure}

The choice of LLMs plays a crucial role in the framework's performance. 
We tested various LLMs to determine their impact on the efficiency of the multi-agent system, as shown in Figure \ref{fig:model_kernel_performance}. 
While our framework utilizes the GLM-4v-9B, we fine-tuned it specifically for our application and compared the performance of our fine-tuned model against other LLMs to highlight the necessity of this fine-tuning.
Fine-tuning allows the model to better understand and adapt to the unique challenges and requirements of our multi-agent autonomous driving scenarios, leading to more consistent and reliable merging. 
%This highlights the importance of model customization to achieve optimal performance in specialized tasks.

%Figure \ref{fig:model_kernel_performance} shows the comparison of merging success rates among different LLMs. 
%Our fine-tuned GLM-4v-9B (Ours) outperforms the other models (LLaVA-NeXT-7B, GLM-4v-9B, and InternVL2-8B) that were not fine-tuned specifically for this application. 
%In medium mode, for example, our model achieves a merging success rate of 0.81, compared to 0.58 for LLaVA-NeXT-7B, 0.65 for the non-fine-tuned GLM-4v-9B, and 0.71 for InternVL2-8B. 
%This demonstrates the necessity and effectiveness of the fine-tuned GLM-4v-9B applied for our framework. 

\paragraph{Conclusion}
The dynamic adjustment of cooperation and efficiency factors, coupled with the RAG mechanism and LLMs, significantly enhances the overall performance of the framework. 
The adaptability provided by LLMs ensures that the CCMA can respond to varying traffic conditions and optimize multiple performance indicators simultaneously. 
%This leads to higher merging success rates, reduced collision rates, lower average merging times, and increased throughput, confirming the effectiveness of our approach in achieving superior performance in complex multi-agent autonomous driving scenarios. 

% Please add the following required packages to your document preamble:
% \usepackage{multirow}

\begin{figure}[htbp]
\centering
\includegraphics[width=0.45\textwidth]{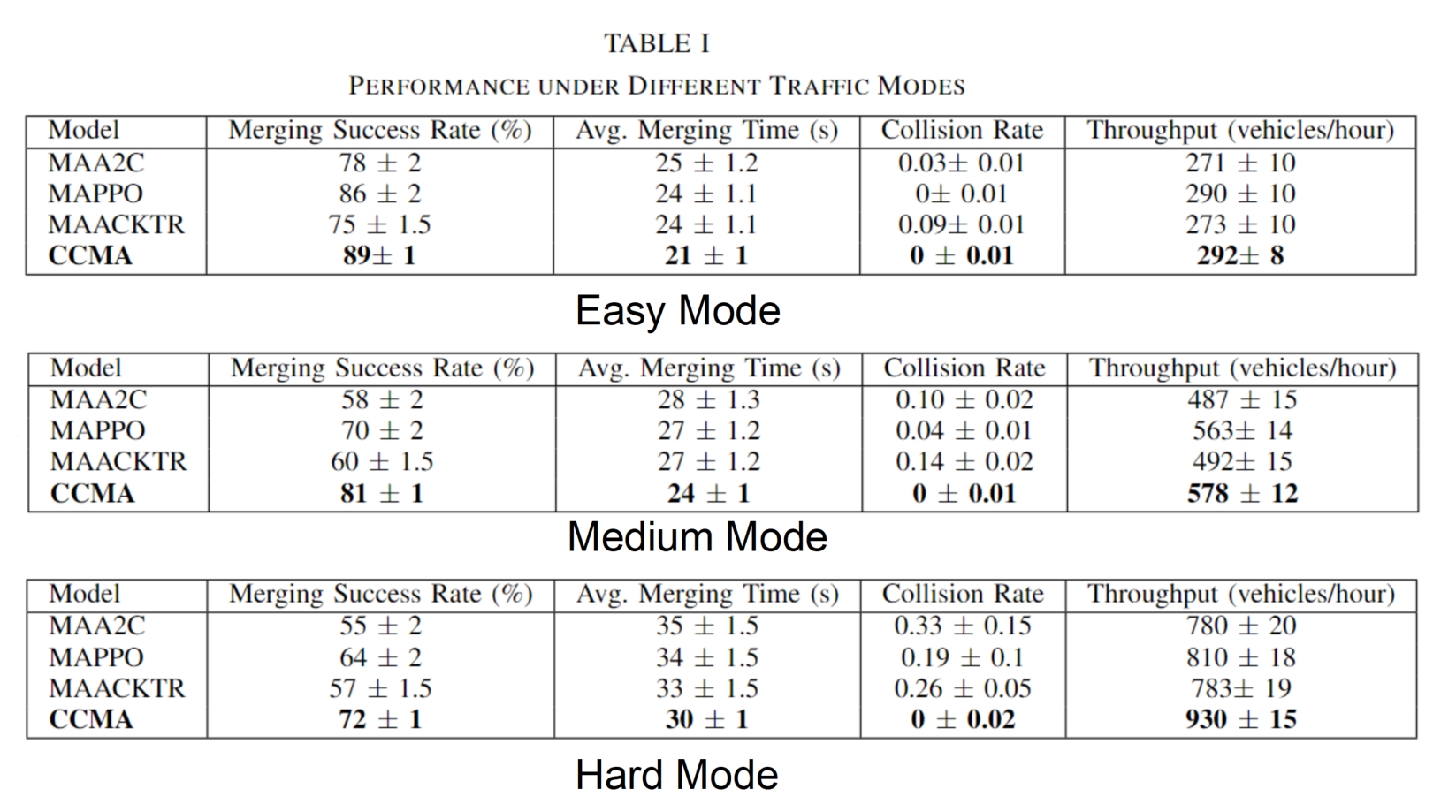}
\label{fig:table}
\end{figure}

\subsection{Overall Results}
The performance of our CCMA framework is summarized in Table I.
In low-density traffic conditions, the CCMA achieved a 0.89 merging success rate with an average merging time of 21 seconds and no accidents. This suggests that the CCMA operates efficiently and safely in light traffic.
Under medium-density traffic, the merging success rate of the CCMA dropped to 0.81, and the average merging time increased to 24 seconds. While still effective, the system faces slightly higher complexity and risk in moderate traffic.
In high-density traffic, the CCMA maintained a 0.72 success rate with an average merging time of 30 seconds and no accidents. These results underscore the need for improved algorithms to handle heavy traffic more effectively.

\subsection{Discussion}
By integrating LLMs with a structured communication protocol, our approach facilitates more nuanced and dynamic interactions between autonomous agents, proving crucial for real-world applications such as intelligent transportation systems. 
The inclusion of additional metrics such as throughput and efficiency provides a more comprehensive evaluation of system performance, highlighting areas where further improvements can be made. 
Future work will focus on enhancing the coordination mechanisms and exploring the impact of different traffic scenarios on system performance.

\section{Conclusion}
\label{sec:conclusion}
This study presents the Cascade Cooperative Multi-agent (CCMA) framework, aimed at enhancing autonomous driving behaviors through the integration of RL models and LLMs. 
The CCMA framework innovatively streamlines multi-agent optimization by segmenting it into individual, regional, and global levels. 
RL is utilized for individual-level optimization, while LLMs facilitate coordination at the regional and global levels. 
Experimental results indicate that the CCMA framework significantly outperforms traditional algorithms, achieving higher success rates, lower collision rates, and enhanced traffic efficiency. 
Notably, the incorporation of regional driving styles, the RAG mechanism, and optimized reward functions facilitate dynamic optimization and continuous system evolution. 
In all, our CCMA framework allows for more efficient and safe interactions among autonomous agents, paving the way for advancements in autonomous driving technology.

\addtolength{\textheight}{0cm}   % This command serves to balance the column lengths
                                  % on the last page of the document manually. It shortens
                                  % the textheight of the last page by a suitable amount.
                                  % This command does not take effect until the next page
                                  % so it should come on the page before the last. Make
                                  % sure that you do not shorten the textheight too much.

%%%%%%%%%%%%%%%%%%%%%%%%%%%%%%%%%%%%%%%%%%%%%%%%%%%%%%%%%%%%%%%%%%%%%%%%%%%%%%%%

%%%%%%%%%%%%%%%%%%%%%%%%%%%%%%%%%%%%%%%%%%%%%%%%%%%%%%%%%%%%%%%%%%%%%%%%%%%%%%%%

%%%%%%%%%%%%%%%%%%%%%%%%%%%%%%%%%%%%%%%%%%%%%%%%%%%%%%%%%%%%%%%%%%%%%%%%%%%%%%%%
% \section*{APPENDIX}

% Appendixes should appear before the acknowledgment.

% \section*{ACKNOWLEDGMENT}

% The preferred spelling of the word ÒacknowledgmentÓ in America is without an ÒeÓ after the ÒgÓ. Avoid the stilted expression, ÒOne of us (R. B. G.) thanks . . .Ó  Instead, try ÒR. B. G. thanksÓ. Put sponsor acknowledgments in the unnumbered footnote on the first page.

%%%%%%%%%%%%%%%%%%%%%%%%%%%%%%%%%%%%%%%%%%%%%%%%%%%%%%%%%%%%%%%%%%%%%%%%%%%%%%%%

% References are important to the reader; therefore, each citation must be complete and correct. If at all possible, references should be commonly available publications.

\end{document}